\documentclass[conference]{IEEEtran}
\IEEEoverridecommandlockouts
\usepackage{cite}
\usepackage{amsmath,amssymb,amsfonts}
\usepackage{float}
\usepackage{placeins}
\usepackage{algorithmic}
\usepackage{graphicx}
\usepackage{textcomp}
\usepackage{xcolor}

\begin{document}

\title{A Surveillance Based Interactive Robot\\
}

\author{\IEEEauthorblockN{Kshitij Kavimandan}
\IEEEauthorblockA{\textit{Department of Computer Engineering} \\
\textit{NMIMS University}\\
Mumbai, India \\
kshitij.kavimandan46@nmims.edu.in}
\and
\IEEEauthorblockN{Pooja Mangal}
\IEEEauthorblockA{\textit{Department of Computer Engineering} \\
\textit{NMIMS University}\\
Mumbai, India \\
pooja.mangal57@nmims.edu.in}
\and
\IEEEauthorblockN{Devanshi Mehta}
\IEEEauthorblockA{\textit{Department of Computer Engineering} \\
\textit{NMIMS University}\\
Mumbai, India \\
devanshi.mehta49@nmims.edu.in}
}

\maketitle

\begin{abstract}
We build a mobile surveillance robot that streams video in real time and responds to speech so a user can monitor and steer it from a phone or browser. The system uses two Raspberry Pi 4 units: a front unit on a differential drive base with camera, mic, and speaker, and a central unit that serves the live feed and runs perception. Video is sent with FFmpeg. Objects in the scene are detected using YOLOv3 to support navigation and event awareness. For voice interaction, we use Python libraries for speech recognition, multilingual translation, and text to speech, so the robot can take spoken commands and read back responses in the requested language. A Kinect RGB-D sensor provides visual input and obstacle cues. In indoor tests the robot detects common objects at interactive frame rates on CPU, recognizes commands reliably, and translates them to actions without manual control. The design relies on off-the-shelf hardware and open software, making it easy to reproduce. We discuss limits and practical extensions, including sensor fusion with ultrasonic range data, GPU acceleration, and adding face and text recognition.
\end{abstract}


\section{Introduction}

One of the important applications of robots is surveillance \cite{b2}. Surveillance is perceiving the environment. This commonly occurs in a military application where monitoring the enemy’s location and the borderlines is important to a country’s safety. The process of deploying humans near sensitive regions is called human surveillance.  This type of surveillance is limited because humans cannot be deployed in risky and inaccessible locations. The great advancements in networks and robot technology provide the feature of monitoring the critical areas remotely using robots instead of humans.

The traditional surveillance systems require continuous monitoring by some dedicated personnel, which is not possible in every household. Hiring an unknown person to do so will also raise privacy issues. There are limitations associated with the CCTV cameras installed these days.

With the development of wireless communication and the internet, security systems are rapidly improving. This paper describes a method for controlling a robot through user commands. The main unit of the robot is a  Raspberry Pi. Using a motor driver IC, two DC motors are connected to the GPIO of the Pi. Microsoft Kinect is the sensor used to detect images and obstacles. A web server is a Raspberry Pi using the FFmpeg streamer program. Through this, the live feed is presented to the user on his/her computer screen. 

The arrangement of this research paper is as follows: Section 2 introduces the literature survey, Section 3 presents the proposed system design, Section 4 illustrates the flow chart of the proposed system's program, Section 5 shows the results and discussion, and finally, Section 6 gives the conclusions. 

\section{Related Work}
The design and implementation of the surveillance tanked robot \cite{b2} was introduced, based on the Wi-Fi protocol and Windows operating system. The movement directions of the robotic tank are controlled by a GUI designed using a Visual Studio development environment. The robot can transmit real-time video to the intended recipient. Also, it can pick and place objects using a (4-DOF) robotic arm.

Staffan Ekvall and Danica Kragic had presented methods toward integrating spatial and semantic information in a service robot scenario \cite{b3} that allows the robot to reason beyond a simple geometrical level. They are interested in using different learning techniques to acquire the semantic structure of the environment automatically. In this paper, this has been demonstrated in an object detection/recognition scenario. They have built a SLAM system using different types of features (lines, points, SIFT) and sensors (laser, camera) and have shown how to construct a navigation graph that allows the robot to find its way through the feature-based map. They have also shown how they can partition this graph into different rooms with a simple strategy. 

Genichiro Kikui introduced three approaches in creating corpora for developing speech-to-speech translation \cite{b4}. They are 1) collecting sentences from reference books compiled by bilingual experts (BTEC), 2) expanding the corpus by paraphrasing, and 3) dialogues with an S2ST system. The first two approaches were intended to cover wider subjects and expressions, while the last approach focused on collecting actual utterances. 

The comparison done by Rami Matarneh is mainly based on accuracy, API, performance, speed in real-time, response time, and compatibility \cite{b5}. The variety of comparison axes allows us to make a detailed description of the differences and similarities, which in turn enables us to adopt a careful decision to choose the appropriate system depending on our needs.

Abhijit Kundu and K Madhava Krishna presented geometry-based techniques \cite{b6} for detecting multiple moving objects and people from a moving single camera. The system was able to successfully detect various moving persons and objects even when they performed degenerate motion, as depicted in experimental results. The system requires a single camera and odometry, which is easily available on most robots. Other sensors like laser and stereo camera can be easily integrated into the system, which can give accurate depth information, and will result in a smaller bound in the FVB constraint. The proposed method is real-time. Their system was able to reliably detect independently moving objects at more than 30 Hz using a standard laptop computer, which is also simultaneously running other routines like obstacle avoidance.  Also, the entire technique uses only grey-level information. Thus, it does not require the person to wear a distinct colour from the background and is more robust than colour-based approaches to lighting changes. The methodology presented here would find immediate applications in various applications of moving objects and person detection, such as in surveillance and human-robot interaction.

A new DNN-based object recognition and robot navigation method \cite{b7} was proposed by Marianus Kurniawan. The objects recognised by the robot are placed in an apartment living room. Two types of images, RGB and Depth images, are used as input to the DNN. The neural network classified each object class with an average accuracy of more than 80\%. The overall training and validation accuracy were 98\% and 97.65\%, respectively. The performance is tested in different object orientations and distances. The developed system is a master-slave one, where the DNN runs on a server because it is computationally expensive. The user sends high-level commands to the robot, like searching for and reaching a specific object, through the server. The robot is controlled by a Raspberry Pi. The robot successfully searched, recognised, and reached the object in real environment experiments. The performance is evaluated in terms of object recognition at varying distances and orientations of the objects. Robot navigation reaching the objects in an arbitrarily selected order shows a good performance of the proposed method. 

A vision-based navigation algorithm was proposed in \cite{b8} by Abhijith R. Puthussery, using machine learning to identify objects in the environment as markers in real-time. Robot Operating System and Object Position Discovery system were used to orient to the marker, calculate the distance and navigate towards these markers using a depth camera. Here, the robot used was the Kobuki TurtleBot 2. It has reliable odometry sensors, a mobile base, an Asus Xtion RGBD camera, and an ODROID XU4 octa-core microcomputer. The proposed method extracts markers from images captured by the robot and navigates to desired targets autonomously. The algorithm for marker detection and robot navigation was successful in achieving the desired results. The custom image recognition engine works well with the developed ROS interface. The experimental results show that the proposed algorithm worked well with good accuracy.

\section{System Design and Architecture}

\subsection{Raspberry Pi} 
Raspberry Pi is a series of mini single-board computers developed by the Raspberry Pi Foundation with the intent to promote the teaching of basic computer science in schools and developing countries. In our context, the Raspberry Pi bears the potential for image processing and speech processing. The use of Raspberry Pi over an actual desktop or laptop PC is chosen to make the robot an embedded system. Python and OpenCV image processing libraries are used for our task.

\begin{figure}[htbp]
\centerline{\includegraphics[width=8cm]{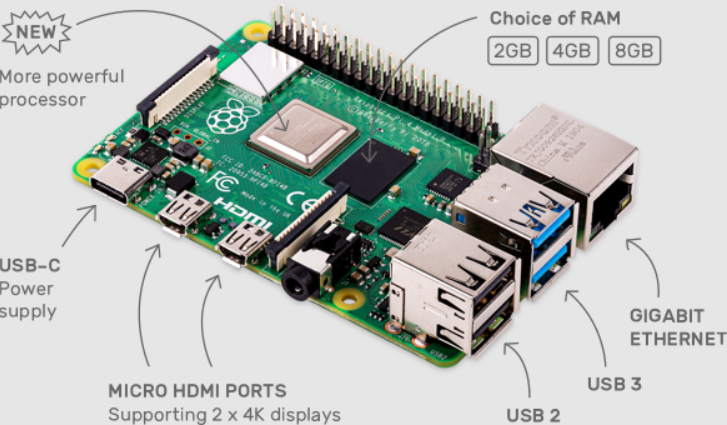}}
\caption{Raspberry Pi 4 module}
\label{fig}
\end{figure}
In the proposed system, two servers, front and central, were made using two Raspberry Pi modules. These were used for communication between the user and the robot. For this purpose, the front server is mounted on a mobile differential robot for manoeuvring using odometry. The front server is connected to a camera, which provides the live feed to the central server in order to let the user monitor the surroundings of the robot. 
\subsection{Object Detection}
To successfully detect surrounding objects, we researched various object detection methods. One of them included the inception v3 model trained on the Imagenet dataset. This method does identify objects, but it is slow. Considering the requirement of real-time and fast object detection, in this project, we use the You Only Look Once (YOLO) model. YOLO efficiently provides an efficient object detection with an extremely fast speed as compared to other algorithms.

\subsubsection{YOLO}  
YOLO is a jointly trained method for object detection and classification for real-time video. Using this method, YOLOv3 is trained simultaneously on the COCO detection dataset and the ImageNet classification dataset. It offers an easy tradeoff between speed and accuracy, and for this reason, it is one of the most efficient algorithms for object detection in real-time. 

YOLO applies a single convolutional neural network to an entire image and divides the image into an S x S grid\cite{b6} and comes up with bounding boxes, which are drawn around images and predicts probabilities for each of these regions for object recognition, object localisation, and object detection. YOLO model divides an image into an S×S grid. Each grid cell predicts B bounding boxes and confidence scores for those boxes. These confidence scores reflect how confident the model is that the box contains an object, and also how accurate it thinks the box is that it predicts. Formally we define confidence as $Pr(Object) * IOU_{pred}^{truth}$ .If no object exists in that cell, the confidence scores should be zero. Otherwise, we want the confidence score to equal the intersection over union (IOU) between the predicted box and the ground truth.

Each bounding box consists of 5 predictions: x, y, w, h, and confidence. The (x, y) coordinates represent the centre of the box relative to the bounds of the grid cell. The width and height are predicted relative to the whole image. Finally, the confidence prediction represents the IOU between the predicted box and any ground truth box.
At test time, we multiply the conditional class probabilities and the individual box confidence predictions, $P_r(Class_i|Object) * P_r(Object)* IOU_{pred}^{truth} = P_r(Class_i) * IOU_{pred}^{truth}$, \cite{b8}\\
which gives us class-specific confidence scores for each box. These scores encode both the probability of that class appearing in the box and how well the predicted box fits the object.

\subsubsection{YOLO ARCHITECTURE}

\begin{figure}[htbp]
\centerline{\includegraphics[width=8cm]{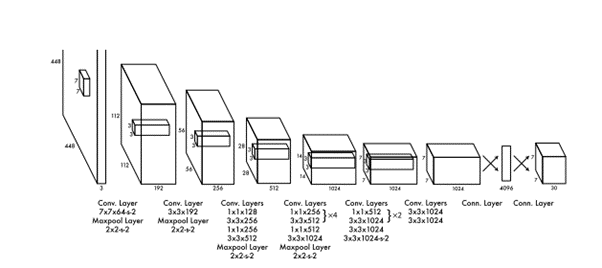}}
\caption{YOLO Architecture}
\label{fig1}
\end{figure}

The network has 24 convolutional layers with 2 fully connected layers. The ConvNet is to extract features from input images, and the fully connected layers are to predict the probability of the boxes' coordinates and confidence score. The accuracies of the predictions also depend on the architecture of the network. The loss function of the final output depends on the x, y, w, h, prediction of classes, and overall probabilities. In our project, we use pre-trained YOLO weights to detect objects. 

\begin{figure}[htbp]
\centerline{\includegraphics[width=8cm]{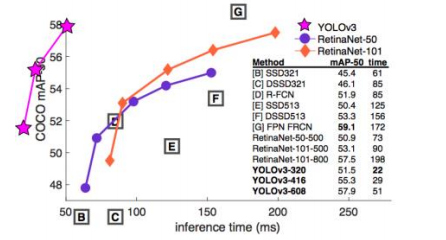}}
\caption{YOLO Performance}
\label{fig2}
\end{figure}
Thus, using the YOLO model, the system recognises and identifies objects of the incoming feed, which helps the robot navigate more efficiently.
\subsection{Speech Recognition}

Speech Recognition is the process of a machine listening to speech and identifying the words. The proposed system recognises what a person is speaking and is also capable of producing a text output of the same. It can also recognise the language of the input speech as well as translate it into the desired language.
Such a system can help users give voice commands to the robot.
Speech Recognition and Translation are done using various libraries in Python, including speech\_recognition, pyttsx3, and googletrans. The speech\_recognition modules first take the input from the user. After getting the input, the Google translator attempts to detect the source language and confirms the language with the user. On getting acknowledgement from the user, the robot asks the language to be translated to. Google Translate then translates our speech into text into the required language. We have used pyttsx3 for text-to-speech conversion. Unlike alternative libraries, it works offline and is compatible with both Python 2 and 3.

\section{Working and Implementation}

\begin{figure}[htbp]
\centerline{\includegraphics[width=8cm]{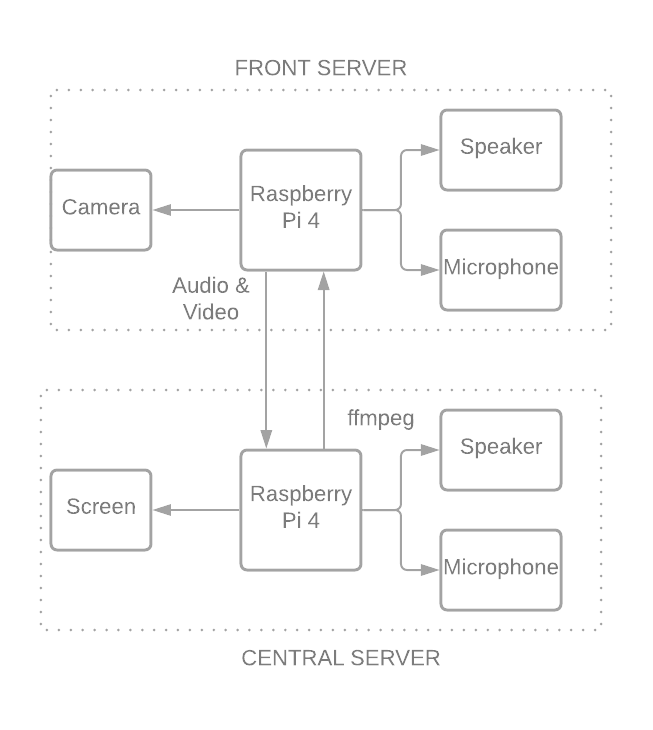}}
\caption{POPEB: Position Oriented \& Perception Enabled Bot}
\label{fig3}
\end{figure}
Two servers were created using two Raspberry Pi 4. The first one, which is the front server, has a camera, a speaker, and a mic connected to it, allowing the bot to take the surrounding inputs. The second Raspberry Pi, which acts as a central server, consists of an output screen that is visible to the user, which again consists of a speaker to give an audio output and a screen for the visual output. There are various steps involved in the integration of the entire process of making the robot. The robot, with the help of the front server, feeds audio and video streams to the central server using the FFmpeg library. On receiving the input, the central server recognises faces, objects, and speech by Object as well as Speech recognition using Machine Learning where the object recognition was done using the YOLO model. The front-end server and central server can be in close proximity or far apart, depending on the situation.

The Robot will be able to recognise multiple languages. It would automatically recognise the language we are talking in and translate it into the required language. This is done by using Google Translate, pyttsx3, and speech\_recognition libraries implemented in Python.

\section{Results and Discussion}
\begin{figure}[htbp]
\centerline{\includegraphics[width=8cm]{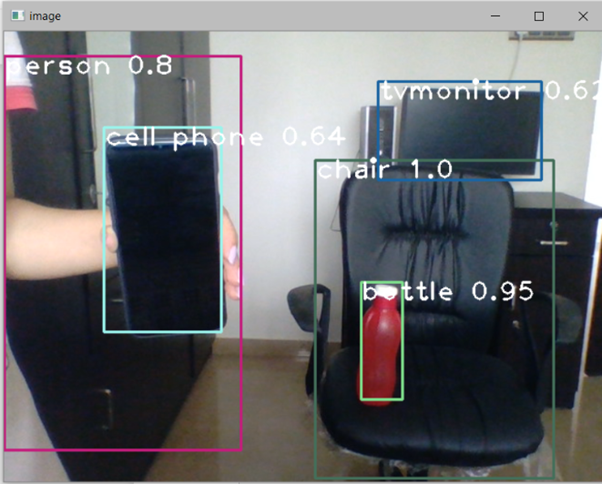}}
\caption{Real-time object detection with multiple bounding boxes}
\label{fig4}
\end{figure}

\begin{figure}[htbp]
\centerline{\includegraphics[width=8cm]{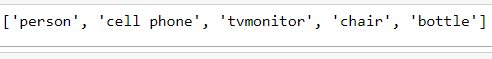}}
\caption{Text Output}
\label{fig5}
\end{figure}

Fig. 5 illustrates the real-time object detection and recognition. The system has successfully recognised every object present in the image based on the trained COCO dataset, along with their confidence of class recognition. Fig. 5 illustrates the text form, which is later converted into speech. The speed of detection and recognition can be increased with the use of a GPU-based system.

\begin{figure}[htbp]
\centerline{\includegraphics[width=8cm]{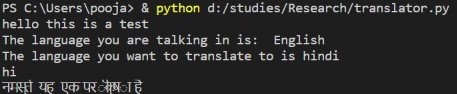}}
\caption{Speech Recognition and Translation}
\label{fig6}
\end{figure}

Fig. 7 shows that speech was recognised and text output for the same was obtained. The language was also recognised and translated into the desired language.
\section{Conclusion}

This work presents the goals and requirements for an interactive robot with the ability to detect objects and recognise speech. The main objective of detecting the desired object using YOLO and Microsoft Kinect RGBD images, and then getting all this information and making a simple navigation control was achieved. Based on this, different approaches were investigated, and YOLO was considered to be well-suited for this project. The system could successfully recognise speech as well as translate it. 

In a nutshell, this work presents a simple robot that can navigate anywhere and provides a live feed to the user who can perceive what is happening in a particular area without being physically present and can also give commands to the robot as in how and where to move. It can also be used as an indoor surveillance system, which could help users monitor their houses/offices in their absence.

\section{Future Work}
In future, we plan on extending the use of this system to outdoor environments as well, by using more efficient recognition algorithms.
We also plan to fusion the Kinect information with other sensors like ultrasonic sensors in order to get a more robust architecture for a mobile robot navigation task. 
Face and text recognition can also be implemented as a future prospect. \\
Another improvement in the future could be to train our own models on different datasets with lower resolutions and see if it could improve accuracy for smaller objects, to help robots recognise objects and navigate more efficiently.

\vspace{12pt}

\end{document}